# Ethical and Scalable Automation: A Governance and Compliance Framework for Business Applications

Haocheng Lin

Department of Computer Science, University College London, UK

Corresponding Author: Haocheng Lin, haocheng.lin.19@ucl.ac.uk

## Abstract

The widespread adoption of Artificial Intelligence (AI) in business operations presents substantial challenges related to ethics, governance, and regulatory compliance. While AI technologies have become integral to daily processes, businesses often lack a cohesive framework for effective risk management. This paper introduces an innovative governance framework designed to ensure that AI applications are ethical, controllable, viable, and desirable. By balancing performance with explainability, the framework helps businesses comply with regulations, such as the General Data Protection Regulation (GDPR) and the European Union (EU) AI Act. Its effectiveness is demonstrated through case studies in the financial and healthcare sectors, where accountability and regulatory compliance are critical. Furthermore, the framework leverages large language models (LLMs, pretrained AI systems designed to process and generate human-like text) to generate synthetic opinions on environmental policies, aligning synthetic outputs with real-world distributions. The selected evaluation metrics, including transparency and performance indicators, enhance explainability without compromising accuracy. Future research will focus on exploring the framework's scalability across industries, such as manufacturing and education, to promote responsible AI integration.

**Keywords:** AI Governance, Ethical AI, AI Compliance, Data Protection, Automation, Large Language Models

## Acknowledgement

This paper builds upon two prior research reports, "AI and Sustainability" and "A Guide to Automating Using AI", published during my research collaboration with Capgemini. I am grateful to Capgemini for supporting these foundational studies, which have contributed significantly to the development of the governance framework presented in this paper.

## Statements and Declaration

Ethics Statement: The research presented in this paper complies with the ethical guidelines for academic research. All data sources were publicly available and anonymised.

Conflict of Interest Declaration: The author declares no conflict of interest in relation to the work presented in this paper.

Funding Statement: The author didn't receive support from any organisation for the submitted work. No funding was received for conducting this study.

# 1 Introduction

## 1.1 Introduction

AI is rapidly transforming society, expanding far beyond business operations to influence sectors, such as healthcare, education, and public services. As AI systems become increasingly integrated into daily life, they have the potential to revolutionise decision-making. However, this growing adoption presents significant ethical, legal, and societal challenges. Malicious actors could misuse AI models by generating biased decisions; and AI might change the balance of power in favour of the minority elite classes. Its vulnerabilities can erode human oversight and displace jobs in traditional industries.

One of the critical challenges AI faces stems from a biased training dataset, which can lead to discriminatory decisions made against marginalised groups. To mitigate these risks, the General Data Protection Regulation (GDPR) and the European Union (EU) AI Act provide more transparent and accountable legal frameworks [1, 2]. In particular, the EU AI Act adopts a risk-based approach to classify AI systems based on how they impact human rights and safety, applying stricter regulations to high-risk applications [2, 3].

Although these regulations are essential for promoting responsible AI deployment, they face challenges in keeping up with the rapid evolution and scaling of AI systems. Current AI technologies could surpass human cognitive and physical capabilities in the next two decades [4]. As AI systems become more autonomous, they risk amplifying existing algorithmic and data biases due to a lack of explainability. This amplification poses significant social risks, including the possible deterioration of democratic processes and a decline in public trust in AI models [5, 6, 7]. These development trends underscore the urgency of introducing robust and adaptable governance models to address the ethical and legal implications of the growing adoption of AI [8, 9].

Moreover, public interaction with AI influences how people perceive these systems and quantifies AI's impact on the society. The public expects AI systems to be more transparent and accountable than traditional systems. However, when AI fails to meet its objectives or produces biased outcomes, the public tends to react more harshly to its mistakes than human errors, eroding trust and acceptance of AI technologies [8, 9]. Maintaining transparent, explainable, and controllable AI systems is critical for fostering public trust and sustaining interest in advancing AI development.

While frameworks, such as AI4People, focus on human-centred ethical principles, this paper helps to introduce real-life use cases for evaluating a distinct practical governing framework while obeying ethical and legal requirements. Unlike the previous designed models, this study uses a framework that incorporates explainability metrics, such as LIME and SHAP, in governance processes, ensuring that AI models remain transparent and scalable across various industries.

## 1.2 Research Objectives

This paper proposes an ethical AI framework, supported by five objectives, to establish a structured approach for integrating ethical principles, regulatory compliance, and practical recommendations into business contexts.

The five research objectives are as follows:

1. Assess the impact of AI-driven automation on business efficiency, productivity, and employee experience.
2. Integrate ethical principles with regulatory compliance into the AI governance framework, focusing on values such as transparency and fairness.
3. Generate actionable recommendations by interpreting AI model results to guide responsible business practices.
4. Evaluate the scalability of AI systems in real-world conditions to determine their effectiveness across different contexts.
5. Leverage explainability and evaluation metrics to communicate model performance, fostering trust and understanding among stakeholders.

# 2 Literature Review

## 2.1 Ethics in AI Automation

Ethical issues are inherently embedded within machine learning models, stemming from an ”original sin” in how these models learn human biases from training datasets [10]. For example, a study conducted by Boston University found that only 15% of respondents correctly identified the surgeon as the child’s mother in the well-known riddle involving a surgeon and a boy [11].

Fairness reflects an AI model’s ability to make decisions without favouring specific stakeholder groups. An extract from “Weapons of Math Destruction” highlighted how AI algorithms can encode historical biases and reinforce existing inequalities [7]. A ProPublica investigation found that COMPAS, an algorithm designed to predict whether a convicted criminal is likely to reoffend, exhibited signs of bias against minorities [12]. However, a detailed analysis of the model’s performance on a sample of 16,000 probationers revealed no evidence of racial prejudice, with the tool achieving an average accuracy of 71%. Potential solutions for improving fairness and accuracy include the introduction of legislation, such as AI4People and the Ethical Guidelines for Trustworthy AI, to address ethical concerns in AI development [13].

While AI4People and the Ethics Guidelines for Trustworthy AI provide foundational guidelines for ethical AI development, they lack the capability to integrate real-time monitoring in dynamic, high-risk sectors such as finance and healthcare. These guidelines remain theoretical and do not offer practical tools to guide implementation in the rapidly evolving systems. This paper proposes a governance framework designed to optimise and integrate ethical principles, enabling businesses to operate in real time with fairness, transparency, and accountability.

Improving accessibility in code development is a major step toward ensuring transparency. A study found that low-code development is becoming a significant trend in project development, with the proportion of low-code developers projected to rise from 20% in 2022 to 40% by 2027 [14]. Low-code platforms enable more individuals to access AI models without the need for professional development teams. However, this raises concerns about maintaining and customising code as projects scale. As low-code projects grow, they could become complex to manage, particularly when the users rely heavily on pre-built utilities. For instance, the built-in functions might originate from third-party applications, posing potential data security risks. Furthermore, as synthetic or low-code technologies become more widely used, the availability of open-source coding materials for training AI models diminishes.

A study estimated 16–25% decrease in interactions from the Stack Overflow forums following the release of ChatGPT [15]. This raises concerns about the availability of training data for AI models in the future. Fewer up-to-date training resources prevent the models from learning a representative coverage of the latest advancements in code development, weakening their ability to generate solutions for recent coding projects, such as building OpenAI LLM APIs or developing React Native mobile applications. Consequently, the increased reliance on AI models may lead to a decline in their problem-solving capabilities, potentially eroding the stakeholders' trust in AI technologies [8].

## 2.2 Legal Frameworks Governing AI

As AI becomes increasingly pervasive, legal frameworks are evolving to regulate its use, particularly in areas such as data protection, risk assessment, and intellectual property rights. Notable examples include the General Data Protection Regulation (GDPR), the Data Protection Impact Assessment (DPIA), and the EU AI Act [1, 2].

The GDPR, implemented in 2018, governs data protection and privacy across the European Union. A key aspect of the GDPR relevant to AI is its emphasis on data protection by design and by default [2, 3]. For example, AI models should process only the minimum amount of data necessary for their purpose, thereby mitigating the risks of excessive or biased data collection. AI systems that manage personal data must comply with the GDPR, ensuring that individuals retain control over their own information. The GDPR also includes the "right to explanation," allowing individuals to demand clarity regarding decisions made by automated systems [3]. This provision identifies the causes behind the generated outputs and ensures accountability by presenting the process transparently to users. This principle supports frameworks such as Data Protection Impact Assessments (DPIAs) for high-risk AI applications, requiring organizations to justify and document the reasoning behind each automated decisions [2, 3].

The provision is particularly relevant for AI applications in hiring, credit scoring, and healthcare, where automated decisions can significantly impact individuals' lives. The DPIA framework offers a structured approach for organizations to identify and manage ethical and legal risks associated with deploying AI systems [3]. This risk assessment ensures that AI systems comply with updated data protection regulations before becoming fully operational.

The EU AI Act complements the GDPR by adopting a risk-based approach to AI governance [1, 16]. The Act classifies AI systems into four categories based on their potential risks to human rights and personal safety: unacceptable, high, limited, and minimal. Strict regulations prohibit the implementation of AI systems classified as posing unacceptable risks, such as those used for mass surveillance in healthcare, law enforcement, or finance [16]. Additionally, high-risk AI systems must comply with requirements for human supervision, ensuring that supervisors possess the necessary training, competence, and authority to effectively oversee AI operations. While the EU AI Act and GDPR establish a foundational legal framework for AI governance by adopting a risk-based approach and reinforcing high-risk management, these regulations fall short in providing actionable metrics for quantifying bias and scalability. The proposed governance framework seeks to address this gap by integrating legal compliance into ethical AI principles, ensuring that businesses have quantifiable criteria to measure fairness and accountability during development. However, this framework must also ensure that AI generated content does not infringe on intellectual property rights.

Traditional intellectual property (IP) laws protect the human creators' contents, but these laws do not cover the complexities of modern AI systems. These AI systems could generate contents inspired by real-world information [16]. Innovations such as Deepfakes illustrate these challenges, as they can undermine trust and create societal division by altering content, such as images, videos, or audio, to misrepresent real-time events. While supporters of Deepfakes justify their use under the principle of "freedom of speech," they can have serious and damaging consequences for society. For example, cases on Reddit where users superimposed individuals onto explicit images have caused significant emotional and psychological distress. As AI systems continue to evolve, legal frameworks must adapt to address inappropriate content and define intellectual property rights for AI-generated works [16].

In 2023, an AI-generated creation winning the Sony World Photography Award sparked a debate about whether people should consider AI-generated art in the same as traditional art [17]. The artist, Boris Eldagsen, used the competition as an opportunity to raise awareness about the future uses of AI-generated images, arguing that they "shouldn't compete with photographic and artistic works" as they are "different entities."

## 2.3 AI Governance and Scalability

As AI systems scale up, it becomes increasingly challenging for policies, processes, and tools to ensure robust governance [9]. To address this, MLOps (Machine Learning Operations) has emerged as a framework for managing the lifecycles of machine learning models. MLOps ensures scalability and compliance with ethical and legal standards by leveraging tools that support model development, testing, and version control. For automating tests, the continuous integration systems automatically evaluate and integrate updates in ML models. This continuous oversight is critical for businesses as their AI systems expand to oversee more complex tasks.

As businesses scale up, people require new frameworks for producing scalable solutions while ensuring that the stakeholders can maintain supervision over the models [9]. While MLOps effectively manage the lifecycles of AI systems, they sometimes fail to address ethical and legal challenges at scale. The proposed framework in this paper incorporates best MLOps practices, quantified through key metrics such as improving model accuracy by at least 5% during each training session, reducing hallucinations in model outputs to within 2% over a six-month development period, and ensuring the solution's reproducibility across different code development environments. These metrics guide the model in complying with accountable and legal practices, promoting consistent performance and alignment with data protection measures.

Although concerns about the scalability of AI systems persist, LLMs have evolved in recent years to process extensive real-life datasets and better understand people's perceptions of political and environmental issues. A study demonstrates this potential by using GPT-4.0-Turbo to generate estimated distributions of vote shares for the 2024 EU Parliamentary Elections in Germany [18]. This highlights the ability to use socio-demographic training data from existing surveys to scale predictions of voting intentions on a national level, despite the models' limitations in predicting unforeseen data distributions.

## 2.4 Gap in Current Research

Despite advancements in ethical AI and legal compliance frameworks, a significant gap remains in integrating these areas, particularly when addressing the scalability of AI systems. Most existing frameworks focus exclusively on either ethical AI or legal compliance, failing to explore how these two domains interact in scalable AI environments [13, 19]. For instance, the Ethics Guidelines for Trustworthy AI emphasize accountability, fairness, and transparency but do not provide explicit guidance on integrating these ethical principles with legal frameworks such as GDPR, which are crucial for sectors like finance, healthcare, and environmental sciences [13].

Similarly, legal frameworks such as GDPR and the EU AI Act primarily focus on compliance with data protection and risk management but often overlook broader ethical issues, such as bias mitigation and fairness [2, 16]. Having a fragmented approach leaves businesses struggling with balancing ethical concerns and legal mandates, particularly as their AI systems scale up [2, 16, 19]. Additionally, adhering to ethical guidelines can be challenging in environments where LLMs occasionally produce hallucinated responses, such as during multiple-choice tasks [20]. Natural language processing techniques can address this issue by interpreting random answers and mapping them to the most similar close-ended response options from the expected answers. However, while LLMs can predict public sentiments, they sometimes overestimate the proportion of people concerned about the impact of climate change.

The performance differences underscore the need for an integrated framework that balances ethical governance with legal compliance while ensuring the representation of responses from diverse stakeholders' perspectives. This gap highlights the importance of future research aimed at developing governance models that simultaneously address these challenges, ensuring that AI systems remain both ethically and legally sound as they scale up in business environments [8, 9].

## 2.5 Accountability in AI Systems

Accountability is a critical pillar of AI governance, ensuring that stakeholders are responsible for shaping the AI system's decision-making. As AI systems become increasingly integrated into high-risk sectors, such as healthcare and finance, it becomes more important to hold them accountable for their actions. Highly automated systems often lack transparency at explaining how they generate their outputs due to their complex architectures, which poses a substantial amount of risks to both ethical governance and ensuring legal compliance.

For instance, GPT-3.5, has 175 billion parameters trained on forty-five terabytes of text, operates in a non-linear manner, meaning that the relationships between inputs and outputs are not predictable. A complex network of neural layers causes the models' non-linearity, where the LLMs learn complex data patterns that are not directly interpretable by humans. Due to the vast number of parameters and complex probability distributions, LLMs lack the transparency required for ensuring accountability since it can be impractical for tracing through different layers of probability distributions in the LLMs. A sample study has highlighted that adding interpretability fosters trustworthy and transparent decision-making, enabling the stakeholders to hold the models and their developers accountable [8].

To further address the accountability gap, the designed governance frameworks must incorporate human oversight, outlined in the ethical guidelines, such as AI4People and European Commission's Ethics Guidelines for Trustworthy AI [13]. Human oversight monitors, adjusts, or overrides existing AI-made decisions. However, it can be challenging to enforce accountability when the developed AI models grow more complex.

While frameworks, such as MLOps, manage the lifecycle of machine learning models, they do not fully address ethical and legal risks associated with AI systems [9]. Real-time monitoring systems that track the AI outputs and provide feedback loops for correcting biases or errors are essential for maintaining accountability in scalable AI operations. For example, a developed continuous integration system tracks the development and testing of the machine learning models and their data during the development cycle. Any fresh updates will be queued into a task queue for the application to assess and integrate any potential updates into the AI systems.

# 3 Methodology

This methodology outlines the development of a unique framework that integrates ethical AI governance with rigorous legal compliance, specifically designed to adapt in scalable, high-risk business environments, such as healthcare and finance. The framework goes beyond existing methods by including real-time monitoring and multidimensional accountability metrics, enabling solutions that address the opinions of stakeholders.

Selected metrics, such as the Chi-Square test, Jaccard Index, and Mutual Information, quantify the differences between the expected and synthetic response distributions, ensuring accountability by enhancing transparency in how models process the inputs. Selected case studies and expert feedback offer valuable insights for balancing between ethical accountability and regulatory mandates, enabling the model to adapt and generate robust solutions in diverse business scenarios.

## 3.1 Research Design

The proposed framework integrates ethical AI principles, such as regulatory compliance, and domain-specific adaptations, structured around three key components: real-time monitoring, multidimensional metrics, and adaptability mechanisms.

**Real-time Monitoring**

The framework uses a dynamic system for continuously tracking and evaluating AI system output. With longitudinal datasets, such as the UKHLS dataset, this framework ensures that people can perform extrapolations for predicting future trends. For instance, environmental policy simulations help to identify the inconsistencies from synthetic responses, which provides the essential feedback for the developers to optimize data collection on guiding the development of environmental policies.

**Multidimensional Metrics**

The framework uses specialized metrics to quantify and enhance model performance. Metrics such as Chi-Square tests, Jaccard Indexes, and Mutual Information Scores enable a robust evaluation of the synthetic and real-world response distributions. These metrics guide iterative improvements made to ensure transparency and accountability (see Section 4 Framework Development).

**Adaptability Mechanisms**

Expert feedback and case studies help to ensure an adaptable framework that dynamically responds to different domain-specific challenges. For instance:

- In environmental policymaking, adaptability mechanisms help refine predictions as socio-demographic variables evolve.
- In finance, adaptability ensures that the datasets remain aligned with the fluctuating economic indicators, which ensures that the models could provide accurate and actionable recommendations.

## 3.2 Data Sources

### Data Overview

The case studies focus on using real-world datasets from sectors, such as healthcare, finance, and law enforcement. The sources provide insights into how AI governance frameworks are integrated into regulatory frameworks, such as data protection (GDPR), risk management (DPIA), and transparency regulations (EU AI Act) [1, 2]. Specific datasets, including the UK Household Longitudinal Study and SustainBench, offer socio-demographic variables and sustainability metrics that are essential for understanding the stakeholders' opinions.

The UK Household Longitudinal Study dataset comprises of longitudinal responses collected from a sample of 40,000 UK households, surveyed across thirteen waves of studies conducted between January 2009 and May 2023.

It includes socio-demographic variables, such as age, highest qualification obtained, ethnicity, and current job. These variables were from a review of recent studies, including Whitmarsh's report, which highlights factors influencing the stakeholders' opinions on environmental issues, such as household size, political attitudes, and family [21].

### Data preparation and conditioning

A series of preprocessing steps prepare the UKHLS data for conditioning and training selected LLMs. Imputation replaces the invalid values, such as -8 (inapplicable), -2 (refusal), and -1 (do not know), with estimated distributions, ensuring that the data remains representative of the stakeholder population. Additionally, the data distribution is standardised into probability distributions to ensure that all features contribute proportionately to the designed models.

SustainBench, as another example, uses a predefined data loader to automatically preprocess and load the relevant profiling data for tasks, such as poverty estimation, crop estimation, brick kiln classification, and biodiversity assessment. For instance, when analysing datasets that track changes in poverty over time, a custom data loader, poverty_change_dataset, processes inputs such as 255×255×8-pixel satellite images, geocoordinates, survey year, and the number of observations made [22]. As an open-source database, human users maintain SustainBench and ensure its relevance for the tasks at hand.

### Data Selection Criteria and Expert Feedback

To ensure the practical applicability of the proposed AI governance framework, a multi-step approach integrates data selection criteria with insights from domain experts. Data selection prioritizes regulatory compliance and representativeness, ensuring that the collected data aligns with GDPR and the EU AI Act while accurately reflecting stakeholder populations. A diverse panel of experts, including legal specialists, oversees the regulatory alignment and AI practitioners provide technical advice through an iterative feedback loop of semi-structured interviews. These experts guide data quality maintenance and model conditioning to ensure adherence to ethical and regulatory standards [8, 23].

Data Selection Criteria To support the development of a robust AI governance framework, data collection incorporates diverse sources such as national surveys and social media platforms. The following criteria ensure that the collected data aligns with ethical and regulatory standards:

1. **Degree of autonomy:** The required data types depend on the level of autonomy AI models need to achieve precision, contextual understanding, and adaptability across tasks. For example, low-autonomy tasks may require structured datasets that are easier for stakeholders to interpret in environments with clear and predefined instructions. A scale from 1 to 5 quantifies autonomy, where one denotes fully manual tasks and 5 signifies automated tasks with minimal human intervention.
2. **Stakeholder representation:** The data distribution must represent different stakeholder groups. Quantifying key demographic variables, such as age, income, and education level, is essential for helping developers assess the models' margin of error. Heatmap visualizations highlight the relationships between the profiling variables and the expected responses.
3. **Privacy and data protection compliance:** The dataset must follow privacy and data protection regulations. A binary checklist compares the dataset's conditions against the established criteria for compliance with data protection and privacy standards.
4. **Scalability across training and testing scenarios:** An adaptability score (rated on a 1-to-5 scale) assesses whether the data requires preprocessing for ensuring compatibility with the training and testing environments.

### Expert Feedback

A diverse range of expert perspectives is essential for addressing and mitigating potential errors that could arise from AI models' outputs. Expert feedback helps refine AI models by ensuring that they obey strict regulatory standards, such as those outlined in the GDPR and EU AI Act [2, 16]. The following criteria involve quantifiable metrics to evaluate how businesses implement these principles:

1. Understanding the level of automation adopted by businesses.
2. Measuring compliance with ethical and legal guidelines during model development.
3. Assessing the scalability of AI systems when adapting to different training and testing datasets.

Additional criteria guide the selection of experts based on their knowledge and contributions to this study. These requirements, categorized by role, are as follows:

1. Lawyers and regulatory specialists: Experts in GDPR, the EU AI Act, and intellectual property laws are essential for ensuring that the framework reflects current legal requirements. Their feedback helps refine the framework by introducing checkpoints to protect data and ensure compliance with regulations.

2. Machine learning engineers, data scientists, and AI specialists: These professionals provide insights into the challenges of scaling AI systems and offer guidance on fine-tuning models to ensure their outputs obey ethical principles [19, 24]. Their feedback enables adjustments to improve the models' ability to process large and diverse datasets effectively.
3. Domain-specific experts: These specialists interpret AI model outputs, minimizing the risks of hallucinating responses and validating the results for domain-specific applications. Their insights help mitigate such risks, improving the model evaluation scores in specific business contexts.

A case study highlights the value of human annotators in validating the AI-generated outputs. Although these annotators' scores ranged from 0.54 to 0.76, they were instrumental in identifying eight common errors from the LLMs' outputs, such as incoherent, incomplete, and contradictory answers. This underscores the importance of having rigorous human evaluation for ensuring the quality of AI models' outputs [23]. The case study also raises concerns about the reliability of social media ratings used in data profiling, as bot accounts or multiple profiles could manipulate the ratings. Implementing account validation and filter algorithms is an effective approach to quality control, ensuring the selection of relevant evaluation profiles.

## 3.3 Framework

This framework addresses the ethical, legal, and operational challenges associated with implementing AI in high-stakes industries, such as education and finance. For each industry-specific application, this framework provides an approach to facilitate responsible and sustainable AI development while ensuring an appropriate balance of privacy and transparency.

**Education**

The application of AI in education offers significant opportunities to enhance productivity and personalize teaching methods. However, it also presents ethical and regulatory challenges, such as ensuring data privacy and mitigating the risks of over-reliance on automated support. This framework employs real-time monitoring to track AI model performance over time, enforcing accountability and ensuring compliance with data protection standards.

A primary focus is the early identification of learning barriers. AI can analyse student performance data to predict potential challenges in learning, enabling timely and tailored interventions, such as extracurricular support sessions and personalized revision plans. For example, a study demonstrated that personalized revision messages improved the performance of undergraduate STEM students [25]. In a randomized controlled trial, students who received targeted messages were more likely to pass compared to those from the control group, with a statistically significant result ($p = 0.0352$).

Utilizing multi-dimensional metrics, such as Mutual Information and Jaccard Index, ensures that personalized support aligns with individual learning needs without infringing on privacy. Additionally, the framework facilitates a transition in the education system from traditional memory-based testing to curriculum designs that prioritize problem-solving in domain-specific scenarios. By leveraging past performance data, this framework optimizes education, promotes transparency and accountability, and ensures that AI-driven educational tools uphold ethical and regulatory responsibilities.

**Finance**

Integrating AI into finance, particularly in supply chain management, offers transformative opportunities to enhance transparency, sustainability, and operational efficiency. However, these benefits come with ethical and regulatory challenges, such as environmental impacts and inadequate resource management. This framework leverages multi-dimensional accountability metrics to assist organizations in achieving their sustainability goals.

One significant challenge in finance is the lack of effective environmental management systems within supply chains. A study by Harvard Business Review found that the supply chains struggle to account for energy use and resource optimization, hindering their ability to meet sustainability goals [26]. AI-driven predictive models can track environmental metrics and provide actionable insights. For instance, the companies could use correlation analysis for quantifying the relationships between operational processes and environmental impact, identifying areas for resource optimization.

Achieving a successful transformation in supply chains requires a multi-dimensional model that evaluates how AI influences planning, coordination, production, procurement, and marketing [27]. This framework is based on findings presented in a prior study, which embedded ethical principles into each stage of the AI transformation process [28].

This framework guides developers in effectively balancing trade-offs among the energy consumption, cost-efficiency, and long-term revenue goals. In Table 1, key implications derived from the findings helps to illustrate how this framework can restructure a sustainable supply chain:

1. Governance processes must enable stakeholders to understand AI's impact and foster trust in a transparent environment.

2. Organizational success relies on cultivating an ethical, AI-centric culture. This framework provides the tools for organizations to adapt their approaches and address AI challenges effectively.

3. Optimizing AI models: This framework ensures that AI models remain aligned with ethical and operational goals.

4. High-quality training data: Collecting and preprocessing high-quality training data is essential for conditioning AI models to make sustainable decisions that align with stakeholder objectives.

Table 1 A table recording the implications of using AI to transform sustainability initiatives [28].

| Dimension | Implications |
| --- | --- |
| Strategy | 1. Sustainability requires “reimaging corporate strategy by creating new modes of differentiation, embedding societal value in products and services, reimaging business models for sustainability, managing new measure of performance, and reshaping business ecosystems” |
| Supply Chain | 1. The use of AI and modern technologies has already impacted the supply chain, including “demand-forecasting models, end-to-end transparency, integrated business planning, dynamic planning optimization, and automation of the physical flow – all of which build on prediction models and correlation analysis to better understand causes and effects in supply chains.” |
| Governance | 1. The implementation of AI requires appropriate ethical AI governance<br>2. Both AI and sustainability require changes to key controls and governance mechanisms<br>3. Both AI and sustainability to require mature governance of organisational trade-offs (like agility vs sustainability or efficiency vs sustainability). Just as organisations accumulate technical debt because of trade-offs about technology choices, so there is a likelihood that organisations will also accumulate and need to manage “sustainability debt.” |
| Processes | Both AI and sustainability require processes to be reimagined:<br>1. AI: “To capture the full promise of AI, however, companies must reimagine their business models and the way work gets done.”<br>2. Sustainability: “Companies will respond to the ever-louder calls by investors and stakeholders for more disclosure and higher-quality, dependable ESG data and reporting. But that alone is insufficient to bring the worlds of strategy and sustainability together and secure resilience and durable competitive advantage while also increasing environmental and societal benefits. The continuous practice of sustainable business model innovation is the engine to do so.” These changes have major impacts on processes. |
| Measurement | 1. Implementing AI to support sustainability requires a comprehensive approach to the measurement of the ethics of AI.<br>2. But measuring the impact of AI is challenging: “Measurement of the environmental impacts of AI compute and applications is limited by a lack of common terminology, recognised standards, consistent indicators and metrics, and varying or optional reporting requirements” |
| Culture and Skills | 1. Sustainability: “This approach is only achievable if an organisation has the right people and sustainability mindset. Empathy, openness, collaboration, and trust helps to achieve the organisational goals.”<br>2. AI: The implementation of AI at scale needs a wide range of skills. These include the innovative technology and data science skills, but also the new skills required across the organisation work in new ways with AI technologies.<br>3. Environmental intelligence: “Environmental Intelligence is a new field of knowledge that exploits the explosion in Environmental data and the rapid advances in Artificial Intelligence to create solutions to some of the most important challenges facing society today.” |
| Data | 1. Waste: The Economist reports that 'between 70% and 90% of data that organisations collect is "dark data" that incurs unnecessary energy costs to transmit and store without being turned into insights and business opportunities'<br>2. Sustainability: Measuring and managing sustainability requires access to new data across the supply chain and within an organisation<br>3. AI: Implementation of AI at scale requires an enterprise-wide implementation of both data and ethical governance. |
| Technology | Gartner have categorised the components of sustainable technology. These are:<br>1. Sustainability by design for new systems (in other words, treating sustainability in the same way as security and other attributes of technology<br>2. Energy-efficient software<br>3. Sustainable data centres of cloud services<br>4. End-to-end design thinking for sustainability<br>5. Energy-efficient hardware and circular economy practices<br>6. Energy-efficient architecture and networking<br>7. Low-carbon energy sources |

## 3.4 Ethical Considerations

This research addresses ethical considerations critical for responsible AI governance in high-stakes applications. Given the sensitive nature of the data and the complexities involved in regulating AI models, it is essential for these systems to obey ethical principles grounded in ethical theories.

**Deontology**

Deontology shapes the AI models' compliance with rules, duties, and obligations. People's actions and their consequences shape its principles. In an AI governing model, custom frameworks define the deontology principles. To comply with GDPR, personal data must be anonymized after removing all identifier information for privacy protection. The users must consent to their data usage before data collection begins, ensuring that they understand the purpose of the research. While the AI systems must protect sensitive or proprietary information, maintaining transparency is crucial for developers to understand data processing and how it contributes to producing an accurate set of outputs [2, 3]. Defining clear boundaries for GDPR implementation on an international scale poses new challenges, as AI can influence public opinions on a large scale, potentially impacting democratic processes. However, a strict compliance with the GDPR regulations might conflict with the need for a robust and comprehensive dataset in conditioning accurate AI models.

**Care Ethics**

Care ethics measure accountability from those affected by the AI models. Using context, empathy, and the stakeholders' interdependence, this theory supports the development of an inclusive AI model to support a diverse set of perspectives.

For example, building trust and gaining respect from stakeholders by using more open-ended questions in simulating LLMs can provide more relevant and diverse information in domain-specific research. Furthermore, contractualism emphasises the importance of mutual agreements with a framework for enhancing the users' trust in the AI systems.

Interactions with the users, such as providing non-disclosure agreements (NDAs), are essential for maintaining confidentiality and preventing the misuse of sensitive information. NDAs protect private AI systems from unauthorised uses, therefore supporting stakeholders' competitive and ethical interests [8]. Balancing privacy and transparency are vital for enabling the stakeholders to understand the methods used in data collection and results evaluation.

**Virtue Ethics**

Virtue ethics focus on morality, by enabling the stakeholders to understand how the models process the data for decision-making. The stakeholders should also be able to verify outputs using explainable techniques, like feature permutation importance, Local Interpretable Model-Agnostic Explanations (LIME), and SHapley Additive exPlanations (SHAP), to ensure that AI models remain interpretable and trustworthy, while remaining aligned with the ethical standards.

It is important to minimise potential bias in data collection by incorporating perspectives from a diverse range of industries and ensuring that the interview questions have a set of open-ended response options [6, 8]. However, new challenges arose when the LLMs generated excessively open-ended responses. For example, in generating responses about an average US voter, LLM-generated outputs averaged 7.78 words, exceeding the 4-word limit [20]. This highlights how prompt engineering can optimize the model outputs and produce relevant and concise information. To reduce the likelihood of random responses occurring, consequentialism evaluates actions based on their outcomes. This evaluation balances its benefits with potential harms as demonstrated from an example about low-code platforms. While low-code development popularises AI development, there are risks of data vulnerabilities and reduced availability of open-source materials. Consequentialism helps with developing governance frameworks with robust security measures while using developers for the maintenance of low-code libraries.

# 4 Framework Development

The four pillars of ethics, control, viability, and desirability supports an ethical AI framework, as illustrated in Fig. 1 from one of the previous researches [29]. These pillars guide the integration of AI in businesses, ensuring that implementations align with ethical standards, regulatory requirements, and organisational goals.

| Ethical<br>Is the proposed automation ethical? Does it meet the ethical principles and policies in place? | Controllable<br>Is the proposed automation controllable? Can it be controlled fast enough? |
|---|---|
| Desirable<br>Is there a business case? Are the risks and potential impacts understood and acceptable? | Viable<br>Is the automation viable? Can the desired results be delivered? Are the required enablers in place? |

**Fig. 1** A visual representation of the four essential principles behind AI-driven automation [29].

Each pillar provides actionable insights to inform the development, implementation, and monitoring of AI systems, promoting responsible and efficient use of technology within organisations.

## 4.1 Four Pillars of Ethical AI

**Ethics**

Ethical AI focuses on assessing whether AI systems comply with guiding principles and operate fairly, transparently, and accountably. For example, in recruitment, an ethical AI model should provide equitable opportunities across diverse applicant pools [2, 7]. Such a model is successful when its outputs are unbiased, uphold individual rights, and align with social justice goals.

**Control**

Controllable AI must enable human supervision and control, ensuring that the automated systems does not cause any unintentional consequences. In healthcare, for example, a diagnostic AI tool should allow the medical professionals to oversee and override any automated decisions, aligning with the EU AI Act's risk-based approach [1, 16]. This pillar reinforces the principle that human judgement is essential to minimising the risks of unexpected outcomes.

**Viability**

Viable AI evaluates AI from a practical perspective, focusing on whether it achieves the desired outcomes without causing unforeseen risks. For example, in planning, an AI-driven system optimizes delivery routes while balancing speed and resource usage without overburdening the company's resources. This pillar ensures that AI deployments are practical, scalable, and aligned with long-term objectives [24, 16].

**Desirability**

Desirable AI evaluates how AI systems minimise risks while maximising benefits. In customer service, a desirable AI system might focus on improving customer satisfaction by ensuring that the interactions with the users are respectful and fulfilling. By emphasising utility and ethics, this pillar fosters AI adoption, supports sustainable business growth, and builds trust with stakeholders [5, 19].

## 4.2 Core Framework Components

Implementing an ethical AI model requires an incremental and transformative approach that incorporates five ethical concepts: governance, analytics, implementation, ethics enablers, and transformation. By complying with the GDPR and the Ethics Guidelines for Trustworthy AI, this framework seeks to ensure fairness, transparency, accountability, and impartiality [13]. A case study using LLMs for analysing public attitudes on environmental policies supports ethical standards during model selection and testing.

This scenario represents a high-risk AI application due to its potential impact on public opinion and future green policies development. Misrepresenting environmental policies could exacerbate the consequences of climate change, such as rising temperatures and sea levels. The data collection and preprocessing stages comply with the GDPR regulations by sampling data from the UK Household Longitudinal Study. The selection of profiling variables enables LLMs to build stakeholder profiles, helping to understand how these profiles influence public attitudes toward environmental issues [2].

A robust set of evaluation metrics ensures accountability by quantifying the performance of AI models to assess their decision-making processes [13]. Examples of these metrics include Chi-squared test scores (Eq. 1), normalized mutual information (Eqs. 2 and 3), and Jaccard indexes (Eq. 4). These metrics measure the similarities between synthetic and expected response distributions.

$$\chi^2 = \sum \frac{(O_i - E_i)^2}{E_i} \tag{1}$$

$$\text{NMI}(X, Y) = \frac{2 \cdot I(X, Y)}{H(X) + H(Y)} \tag{2}$$

$$I(X;Y) = \int_{\gamma} \int_{\chi} P_{(X,Y)}(x, y) \log\left(\frac{P_{(X,Y)}(x, y)}{P_X(x) P_Y(y)}\right) dx\, dy \tag{3}$$

$$J(A, B) = \frac{|A \cap B|}{|A \cup B|} \tag{4}$$

The Chi-squared test assesses statistical independence between variables to identify anomalies in a model's outputs. For example, in environmental policy simulations, the Chi-squared test verifies whether the model is unbiased across various socio-demographic profiles. Normalized mutual information scores quantify the similarities between two data distributions, providing valuable insights into how the models reflect relationships between variables. Developers use normalized mutual information scores to determine if the models' outputs align with real-world data trends, offering insights into how a model adapts to changing contexts. A Jaccard index measures the degree of overlap between data distributions, ensuring that the data remains relevant in environments with evolving profiling variables. When evaluating questions with binary response options, the Jaccard index is particularly effective at identifying inclusivity in a binary environment, where stakeholders are primarily concerned with maximising overlaps between responses to ensure that the synthetic distributions could emulate real-life response distributions on domain-specific questions.

These metrics continuously monitor how the synthetic opinions evolve with new developments in environmental policymaking and socio-demographic changes. This enables developers to promptly detect and address any potential ethical violations [13]. For instance, applying these metrics allow developers to evaluate the model's suitability for answering questions on topics, such as lifestyle, the impact of individual actions on the environment, and potential environmental risks (Table 2).

**Table 1** A table of evaluation metrics quantifying similarities between the synthetic and expected distributions of responses for ten of the selected questions about environmental issues.

| Question | Chi-Square Test | Jaccard Index | Mutual Information |
|---|---|---|---|
| Describe your lifestyle | 15 | 0.6718 | 0.9057 |
| Personal Impact on Climate | 20 | 0.7953 | 1.0 |
| Willing to Pay | 20 | 0.3824 | 1.0 |
| Personal Change | 20 | 0.5152 | 1.0 |
| Environ. Disaster | 20 | 0.6037 | 1.0 |
| Green Tariff | 12.0 | 0.2690 | 1.0 |
| Pollution | 343.6159 | 0.1005 | 1.0 |
| Environ. Group | 1275.5102 | 0.1887 | 1.0 |
| Climate Change Control | 20 | 0.3795 | 1.0 |
| Climate Change Impact | 30 | 0.6311 | 1.0 |

Integrating these metrics not only enhances transparency but also facilitates prompt feedback to adjust models in response to changes, thereby supporting accountability in AI frameworks such as MLOps. This approach enables businesses to scale up their systems responsibly while maintaining public trust [9, 19]. The feedback obtained from applying the evaluation metrics contributes to mitigating bias in AI models [5, 7]. These metrics help refine the selection of profiling variables, improve the conditioning of the models, and identify areas where LLMs perpetuate historical biases in both their results and training data [6].

## 4.3 Application Across Industries

For instance, in credit scoring, these metrics help ensure that predictions do not unfairly disadvantage certain demographic groups. Using Chi-squared tests and the Jaccard Index for continuous monitoring allows developers to assess how the alignment between synthetic and expected distributions changes over time, enabling the detection of emerging biases in creditworthiness assessments. This approach aims to ensure a fair access to financial resources for all individuals [5, 8]. The metrics explain and quantify performance from a bias-variance perspective.

While bias reflects the differences between expected and predicted distributions, variance captures how spread out the responses are. An example shows a graph illustrating two regimes of training and test errors as the number of training instances increases [30]. The graph highlights bias-variance challenges, where Regime #1 corresponds to underfitting, and Regime #2 represents a well-trained model with acceptable test error.

Controllable AI maintains human oversight over AI systems by balancing it with AI autonomy, ensuring the system's alignment with ethical principles. As illustrated from Fig. 2, distinct levels of control can influence relationships between leadership and technical skills at societal, organizational, and user levels. A previous study emphasised how this dynamic redefines peoples' responsibilities in maintaining this delicate equilibrium [29]. Furthermore, a study by Shneiderman (2022) argued that developers must prioritize creating autonomous models adaptable to various levels of human control, allowing for customization based on users' skill sets, ranging from beginners to AI experts [31].

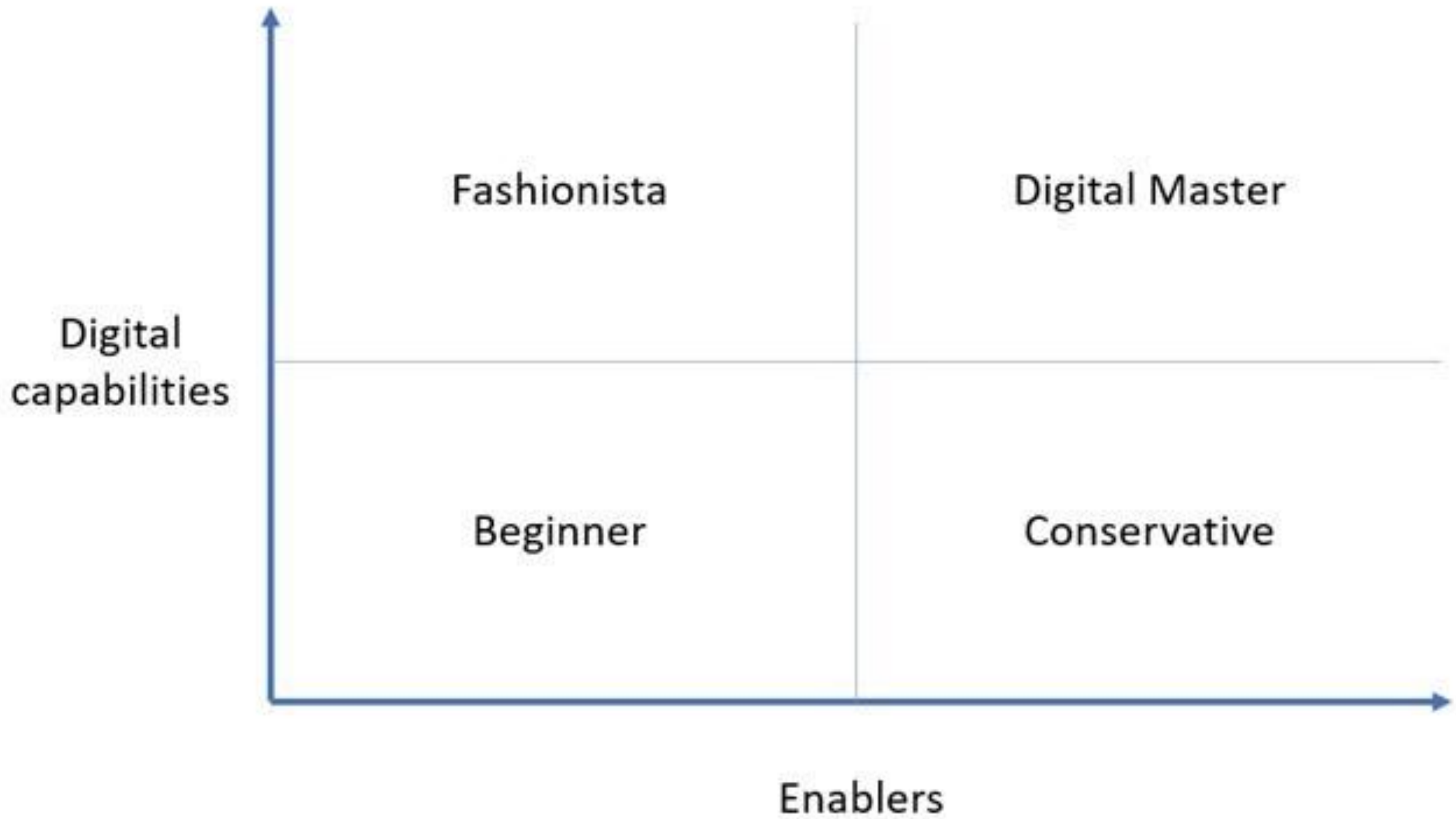

**Fig. 2** A graph showing the diverse levels of digital capabilities against leadership skills [28].

However, the level of human control required depends on the AI models' working environment. In higher-risk environments, human operators must intervene whenever necessary to override any potential risky decisions made by the AI models [1, 16].

For instance, when implementing LLMs to generate opinions on environmental issues, developers must follow a series of steps: preparing the dataset for the models, evaluating whether to rerun and optimise the LLMs after reviewing their results, and determining how to present the findings in a visually appealing manner for the stakeholders to understand.

To maintain control over AI systems as they scale, businesses must continuously monitor and assess them [9, 19]. Fidelity tests ensure that the AI models' data closely resemble real world data while detecting the deviations early. In this paper, dynamic profile conditioning adapts the inputs and outputs to match evolving environments. These functionalities provide a long-term validation strategy to evaluate how models' assumptions and alignments change in response to the latest real-world data.

# 5 Analysis and Discussion

This section provides a critical evaluation of the proposed AI framework and its ability to balance efficiency with ethical principles. Using examples selected from industries such as finance and healthcare, the discussion highlights real-world challenges organizations face when implementing AI frameworks. Industry-specific cases illustrate how this framework addresses trade-offs between operational demands and ethical standards.

## 5.1 Framework Evaluation

The proposed framework integrates key ethical AI principles, such as fairness, transparency, and accountability, with legal frameworks, including the General Data Protection Regulation (GDPR), Data Protection Impact Assessment (DPIA), and the EU AI Act [1, 2, 16]. This integration ensures that AI systems are scalable, efficient, and compliant with legal obligations.

One of this framework's strengths is its ability to address ethical and legal concerns simultaneously. By employing data minimisation principles, models can protect sensitive data [2, 3]. For example, in healthcare, data minimization ensures that the designed AI models could process only essential patient information, such as medical history and current symptoms. One example study shows anonymizing patient data can effectively protect privacy while preserving utility, as demonstrated by the $(k, k^m)-$anonymization algorithm [32]. This approach, from the clustering algorithm pseudocode, anonymizes the records by iteratively generalizing diagnosis codes if they fail to meet the k-anonymity constraints, balancing privacy while minimising utility loss.

The controllability pillar emphasises human oversight in data processing, model training, and testing, ensuring compliance with the EU AI Act's requirement for supervision throughout AI development and deployment [1, 16]. This approach aligns with Shneiderman's advocacy for "high levels of human control" in Human-Centered AI [31]. Shneiderman's work uniquely promotes balancing human oversight with automation by gradually increasing control until the framework meets the required ethical standards.

However, it is challenging to apply consistent AI frameworks across different industrial settings. Various sectors have differing requirements for regulation and risk tolerance. For instance, AI systems in healthcare prioritize transparency and explainability to patients [9]. Regulations also vary by scale, particularly with the popularisation of AI among non-professionals through low-code technology. The control mechanisms used by small teams of individual developers differ significantly from those employed by international business organizations. Additionally, developed AI frameworks may struggle to anticipate unexpected events, as described in the book The Black Swan: The Impact of the Highly Improbable, which outlines the risks of amplified consequences resulting from failures in AI systems [33].

## 5.2 Comparative Analysis

This section provides a comparative analysis of the proposed framework against existing governance models, including AI4People and the Ethics Guidelines for Trustworthy AI. These models establish foundational guidelines for ethical AI, focusing on principles such as fairness, transparency, and accountability. They lack the tools for real-time monitoring and ensuring legal compliance.

The proposed framework distinguishes itself by embedding multidimensional metrics into real-time monitoring. Unlike the AI4People initiative, which emphasises human-centric principles but lacks a set of operational guidelines [13], this framework implements actionable mechanisms, such as Chi-squared tests, Jaccard Index, and Mutual Information, enabling the developers to continuously evaluate the models' performance. A supporting literature review provides the necessary context for interpreting metrics and understanding how these AI models operate in various real-life scenarios. Furthermore, compared to the Ethics Guidelines for Trustworthy AI, this framework could adapt to changes in data patterns and evolving socio-demographic contexts.

This framework incorporates data minimisation and anonymisation techniques to address GDPR requirements while ensuring that data quality is adequate for use in high-stakes environments. Heatmap visualisations illustrate how input features relate to the output (Fig. 3).

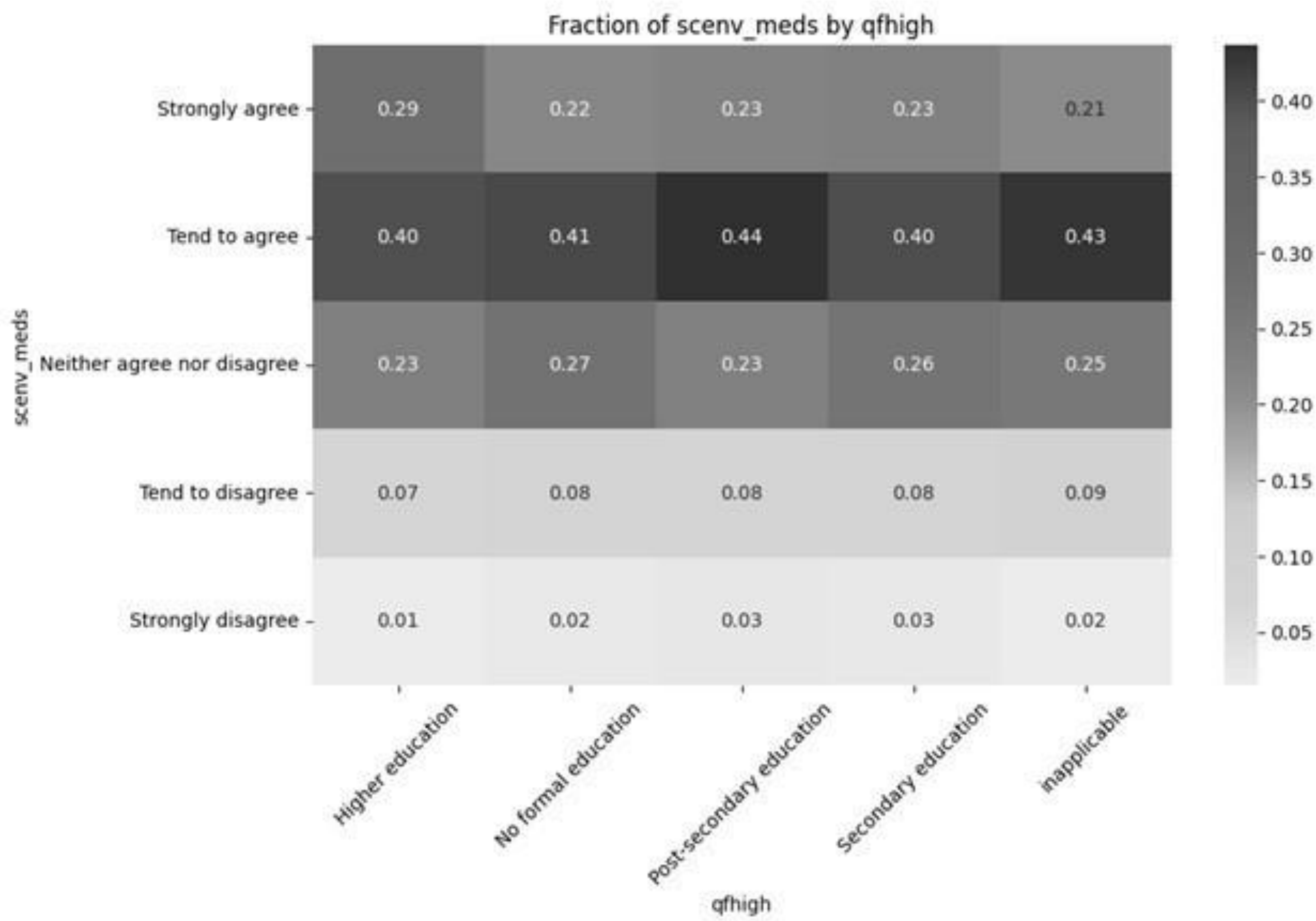

**Fig. 3** An example of a heatmap showing the relationship between the responses to a selected question about whether the world will experience an environmental disaster soon and the corresponding respondents' qualifications.

## 5.3 Trade-offs and Practical Challenges

One of the key challenges in deploying the proposed framework is managing the trade-offs between AI accuracy, fairness, and legal compliance. As AI systems become more complex, achieving high accuracy often requires processing substantial amounts of data, which can conflict with the GDPR's data minimization principle [2]. As datasets grow larger, processing results becomes more computationally complex, as demonstrated in Table 3, which compares the performance of three selected models (Logistic Regression, Random Forest, and MLP) on lowdimensional (Region of Interest, ROI) and high-dimensional (Voxel-Based Morphometry, VBM) data [34]. Table 3 shows that for the same task, the models trained on low-dimensional datasets in an average of 0.62 seconds, while training on high-dimensional datasets took an average of 47.69 seconds. However, after training, both low- and high-dimensional datasets achieved similar prediction times.

**Table 3** Tabular results of the models' performance on low-dimensional (ROI) and high-dimensional (VBM) data.

| Dimensionality | Model | AUC | Balanced Accuracy | Memory Consumption | Training Time | Prediction Time |
|---|---|---|---|---|---|---|
| Low Dimensional | Logistic Regression | 0.84697 | 0.74375 | 872.82011 | 0.09083 | 0.08937 |
| | Random Forest | 0.79924 | 0.7446 | 402.86923 | 0.74577 | 0.0815 |
| | MLP | 0.77273 | 0.72159 | 47.86176 | 1.01017 | 0.0000 |
| High Dimensional | Logistic Regression | 0.83106 | 0.73864 | 1498.64681 | 8.62104 | 0.18906 |
| | Random Forest | 0.78114 | 0.61331 | 1537.54537 | 10.63234 | 0.1764 |
| | MLP | 0.71477 | 0.65682 | 2471.575 | 123.82784 | 0.23906 |

The Receiver Operating Characteristic (ROC) curves from Fig. 4 illustrate that training on lower-dimensional data increases the likelihood of overfitting in the Random Forest (RF) and Multi-Layer Perceptron (MLP) models. These curves help developers compare different models to understand how they balance specificity (false positive cases) and sensitivity (true positive cases). The area under the ROC curve is a vital indicator of model performance, with more accurate models achieving larger areas.

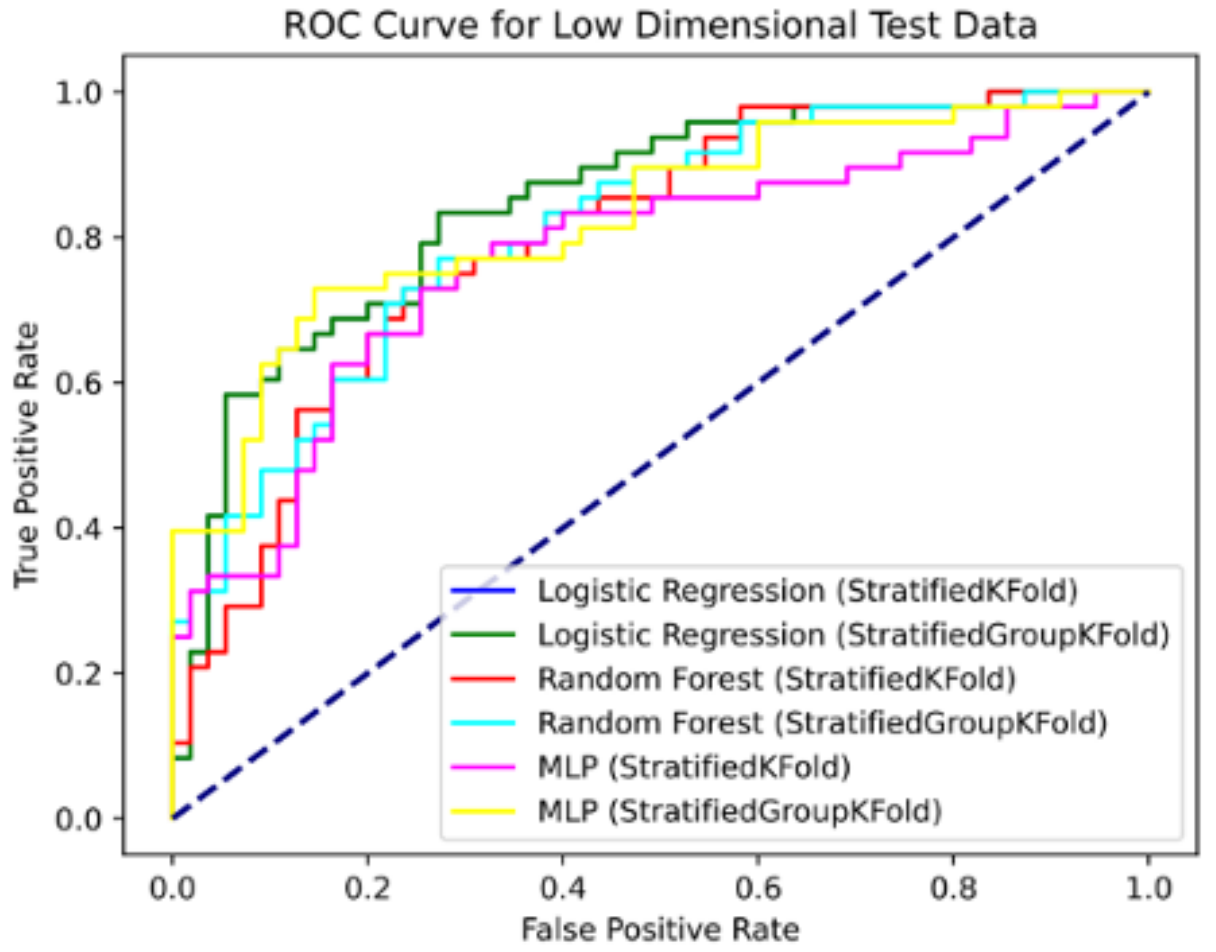

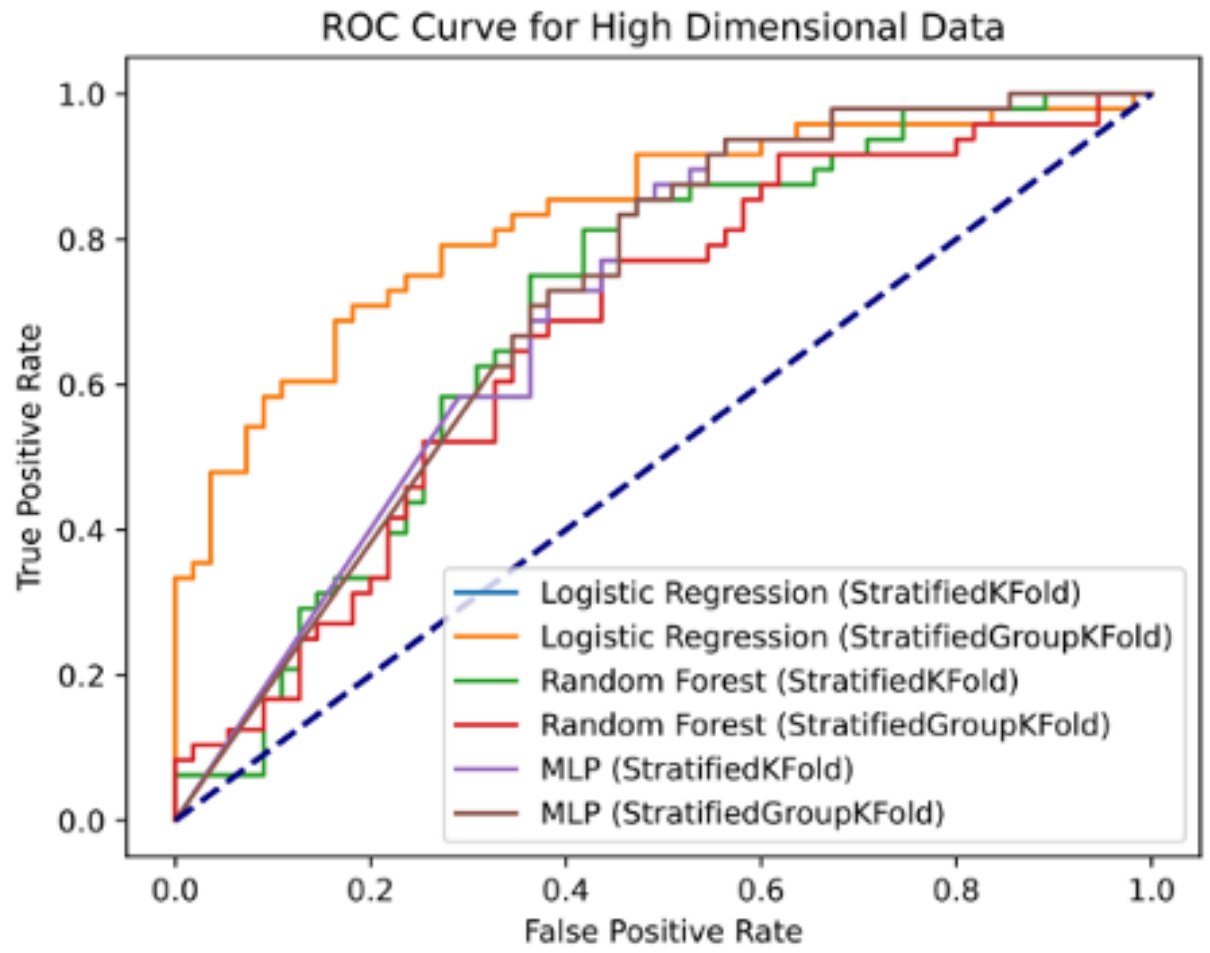

**Fig. 4** Two graphs of comparative ROC analysis on low- (left) and high-dimensional (right) data.

Another challenge lies in balancing data minimisation with gathering enough data to make accurate predictions [2, 7]. Existing AI systems, such as fraud detection models, rely on large datasets for making accurate and adaptable predictions while accounting for evolving fraud patterns. For instance, these models use anomaly detection or behaviour analysis trained on diverse, large datasets to extrapolate fraud patterns [5, 35]. However, the substantial volume of data required to achieve high accuracy can cause conflicts with GDPR's data minimisation principles, as adversarial attacks could exploit different variables for personal identifications, thereby exposing the users to privacy risks.

When developing solutions, businesses must balance innovation with compliance to regulations. For example, Amazon use AI to provide personalised recommendations but sometimes risk violating the data protection regulations. In 2021, the Luxembourg National Commission for Data Protection (CNPD) issued a $886 million fine to Amazon for using personal data to create tailored advertisements for customers [36]. This fine highlights the potential legal challenges faced by companies like Amazon. Feature engineering has become an essential tool for balancing the models' ability to make accurate predictions with data minimization principles. For instance, feature engineering creates, transforms, selects, and optimises features from raw data to enhance the ML models' performance. This involves ranking and selecting variables based on their relevance, as well as generating new synthetic features that better captures the underlying data patterns. Lasso regression is a technique that penalises fewer notable features by reducing their coefficients to zero, effectively discarding the irrelevant data.

# 6 Conclusion

The proposed AI framework represents a significant contribution to AI governance by integrating the principles of ethics, controllability, viability, and desirability. Establishing this four-pillar structure ensures that AI systems comply with the ethical and regulatory requirements. Continuous iterations of monitoring, evaluation, and optimization ensure that the framework has the mechanisms, such as Data Protection Impact Assessments (DPIAs) to align with key regulations, including the GDPR and the EU AI Act [1, 2, 3, 13].

The framework addresses the challenges posed by AI automation, offering solutions for businesses to transform their operations while balancing innovation with legal and ethical obligations. For instance, in healthcare, the framework enables human oversight to ensure accurate diagnoses while emphasizing the principles of ethics, controllability, viability, and desirability. These principles balance with data minimization requirements to protect privacy [2, 13, 16].

Additionally, the designed framework demonstrates its ability to enhance operational efficiency, as evidenced by the reduced task completion times and the alignment between synthetic and expected distributions (Objective 1). Future validation efforts will incorporate additional evaluation metrics to quantify errors and target productivity improvements. Explainable tools, such as SHAP and LIME, are effective at enhancing transparency in the AI models while providing insights into the reasoning behind each model's decision-making process. These evaluation metrics must comply with ethical regulations and promote fair decision-making practices (Objective 2).

Using real-world case studies, this research applies designated metrics to generate actionable feedback, such as optimising profiling variable selections and modifying prompt structures, with the aim of reducing algorithmic biases in treating different stakeholder groups (Objective 3). Past studies, ranging from predicting signs of schizophrenia in local hospitals to understanding attitudes toward environmental policies on a national scale, monitor performance across systems of different scales. This approach minimises errors while ensuring adherence to ethical standards (Objective 4).

Further research should focus on validating the framework across diverse industrial settings to enhance its scalability and adaptability. This effort aims to build trust by fostering a collaborative relationship between AI systems and their human supervisors (Objective 5). For instance, real-time data collection enables AI models to adapt to emerging domain-specific contexts. Additionally, simulations can assess the framework in unique environments to evaluate its reproducibility.

# 7 Data Availability

This study used publicly available datasets to support the findings reported from this article. The datasets used include:

- **UK Household Longitudinal Study (UKHLS):** Longitudinal data spanning Waves 1–13 (2009–2022), available from the UK Data Service (Study Number: 6614). This dataset includes socio-demographic variables critical for understanding stakeholder opinions on environmental policies. Data access is subject to the UK Data Service's terms and conditions. Further details can be found at https://doi.org/10.5255/UKDA-SN-6614-19.

- **SustainBench:** Open-source datasets covering sustainable development goals, such as poverty estimation and environmental quality assessments, preprocessed using publicly accessible data loaders. Details on dataset access and preprocessing guidelines are available at https://doi.org/10.1038/s41467-020-16185-w.

- **Predicting Schizophrenia Dataset:** Edouard Duchesnay, Antoine Grigis, and Benoît Dufumier (Universite Paris Saclay, CEA, NeuroSpin), alongside Francois Caud and Alexandre Gramfort (Universite Paris-Saclay, DATAIA) developed this publicly available dataset.
  This dataset comprises of brain anatomy features collected to facilitate machine learning tasks for predicting schizophrenia diagnoses from RAMP Studio platform:
  https://ramp.studio/events/brain_ anatomy_schizophrenia_UCL_2024 [34].